\pdfoutput=1

\documentclass[11pt]{article}

\usepackage{acl}

\usepackage{times}
\usepackage{latexsym}

\usepackage{mathtools}
\usepackage{mathptmx}
\usepackage{graphicx}
\usepackage[font=footnotesize]{subcaption}
\usepackage[font=footnotesize, justification=centering]{caption}
\usepackage{color}
\usepackage{rotating}
\definecolor{mygreen}{gray}{0.8}
\usepackage{booktabs}
\usepackage{makecell}
\usepackage{xcolor}
\usepackage{tabularx}
\newcommand\ExerciseCaption[1]{%
  \captionsetup{font=footnotesize}%
  \caption{#1}}

\usepackage{booktabs,colortbl}  
\definecolor{rowgray}{gray}{0.9}
\usepackage{amssymb}
\usepackage{pifont}
\newif\ifcaptionlinebreak

\DeclareCaptionOption{linebreak}{\csname captionlinebreak#1\endcsname}

\newcommand\captionlinebreak{\ifcaptionlinebreak\\\else\space\fi}

\DeclareCaptionStyle{mystyle}[linebreak=false]{linebreak=true}
\usepackage[T1]{fontenc}

\usepackage[utf8]{inputenc}

\usepackage{microtype}

\title{DeepParliament: A Legal domain Benchmark \& Dataset for Parliament Bills Prediction}

\author{Ankit Pal \\
  Open Legal AI \\
  \texttt{openlegalai@gmail.com} \\}

\begin{document}
\maketitle
\begin{abstract}
This paper introduces DeepParliament, a legal domain Benchmark Dataset that gathers bill documents and metadata and performs various bill status classification tasks. The proposed dataset text covers a broad range of bills from 1986 to the present and contains richer information on parliament bill content. Data collection, detailed statistics and analyses are provided in the paper. Moreover, we experimented with different types of models ranging from RNN to pretrained and reported the results. We are proposing two new benchmarks: Binary and Multi-Class Bill Status classification. Models developed for bill documents and relevant supportive tasks may assist Members of Parliament (MPs), presidents, and other legal practitioners. It will help review or prioritise bills, thus speeding up the billing process, improving the quality of decisions and reducing the time consumption in both houses. Considering that the foundation of the country's democracy is Parliament and state legislatures, we anticipate that our research will be an essential addition to the Legal NLP community. This work will be the first to present a Parliament bill prediction task. In order to  improve the accessibility of legal AI resources and promote reproducibility, we have made our code and dataset publicly accessible at github.com/monk1337/DeepParliament
\end{abstract}

\section{Introduction}
In recent years, Artificial Intelligence(AI) based methods have been employed in the legal field for several uses and within many sub-areas.
\begin{figure}
    \centering
  \includegraphics[width=5.6 cm]{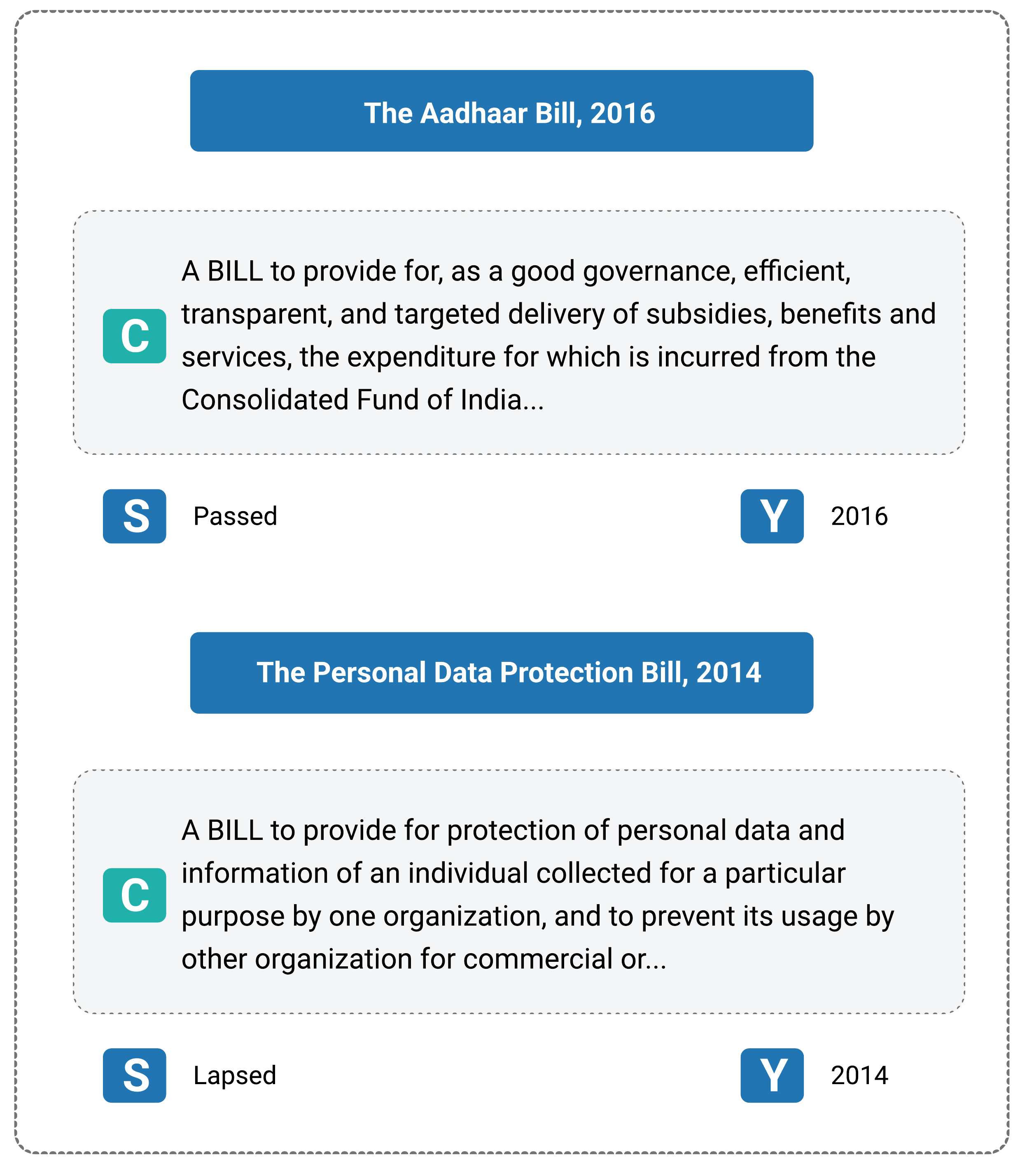} \ExerciseCaption{\footnotesize Samples from the DeepParliament dataset. Here C, S, and Y indicate Bill Context, Bill Status, and Bill Year, respectively.}
  \label{fig:sample_data}
\end{figure}
In the legal field the majority of resources are available in textual format (e.g., contracts, court decisions, patents, legal articles). Therefore considerable efforts have been made at the intersection of Law and Natural Language Processing research. Efforts can be witnessed in the various projects dealing with NLP applications in the legal domain and recently published scientific papers such as legal judgement prediction \cite{Aletras2016, Xiao2018, Chalkidis2019} legal topic classification \cite{Nallapati2008, Chalkidis2020}, overruling prediction \cite{Zheng2021}. Furthermore, researchers have also explored a variety of Legal AI tasks, including legal question answering \cite{Kien2020}, contract understanding \cite{Hendrycks2021}, court opinion generation \cite{Ye2018}, legal information extraction \cite{Chalkidis2018}, legal entity recognition \cite{Leitner2019, Leitner2020} and many more. Predictive legal models have the ability to enhance both the effectiveness of decision-making and the provision of services to individuals. Many new datasets have also  been proposed in the legal domain to track the recent progress and serve as benchmarks. Recently, there have been initiatives to develop corpora for the India's judicial system.

The foundation of India's democracy is the Parliament and state legislatures. Implementing, amending and removing laws is the primary responsibility of Parliament. The Rajya Sabha (Council of States) and the Lok Sabha (House of the People) are the two houses that constitute India's legislature. However, the majority of contemporary legal NLP research focuses only on court decisions and cases. This issue, which we refer to and subsequently characterize as "Parliament Bills Prediction", has not been explored. Before qualifying as an act, every bill passes through a long chain of standardized processes, including introduction of the bill, publication in the gazette, first reading, select committee, second reading, and third reading. After the third reading, the bill goes to the other houses, and after the approval  of both houses it faces final approval by the president. A significant amount of time and effort is required to pass a bill in either of the Houses of Parliament. Therefore, a lapse of a bill has a negative impact on legislative work.

India's first Lok Sabha (1952–1957) passed 333 bills throughout its five-year existence. The average number of bills approved by Lok Sabhas with terms less than three years is 77. Both houses spent about half their time carrying out legislative business. The lapse count of the bill increases at the end of every Lok Sabha. A total of 22 bills lapsed after the 16th Lok Sabha; three bills have been pending for over 20 years; six have been pending between 10-20 years. Legislative activity accounts for a significant portion of Lok Sabha's working hours. To date, 14 bills are still pending  between 5-10 years and 10 bills are pending for under five years. Therefore, time is wasted when bills lapse at the end of the Lok Sabha's tenure, as a new Lok Sabha must start over and consider  bills from scratch, taking at least two sessions to reconsider the bills. Thus, in order to improve productivity, it is necessary to re-evaluate the rule governing the lapsing of bills in the House of Representatives. This calls for machine learning strategies to enhance the efficiency of the billing process in Parliament.

To the best of our knowledge, a single dataset does not yet exist, which provides a standard benchmark for parliamentary Bills. To facilitate research on bill documents for text classification, we provide DeepParliament, a legal domain Benchmark \& Dataset which gathers bill documents and meta data and performs different status classification tasks. The proposed dataset and benchmark are not meant to replace or compete with the decisions of the Houses of Parliament and the President by any means; instead, the proposed solution offers complementing use cases. Models developed for bill documents and relevant supportive tasks may assist members of the Legislative Assembly (MLA), Members of Parliament (MPs), presidents and other legal practitioners, for example by estimating the likelihood of getting a bill passed, reviewing or prioritizing bills, (thus speeding up the billing process); improving the quality of decisions and finally, reducing the time and energy consumption in both houses. Fig. \ref{fig:sample_data} shows two samples of two parliament bills' context, their corresponding status, and the year from the study dataset.

Applications developed on this dataset, such as automatic summaries, would enable the professionals to decide which documents they should read in detail. Moreover, the model can suggest different sections and acts in need of further exploration by highlighting which areas a new bill falls within. This paper proposes a benchmark and takes initial steps by contributing the dataset and baseline models to the community. Moreover, the plan is to continue to revise and upgrade the DeepParliament dataset in the future.

In brief, the contributions of this study are as follows.
\begin{itemize}
    \item \textbf{New Legal Dataset}. We are proposing a new dataset. To our knowledge, there is no dataset focusing on parliament bills and data. Therefore, this work will be the first to present a parliament bill prediction task having rich information on parliament bills, different acts and laws.
    
    \item \textbf{Diversity and Difficulty}. The proposed dataset text covers a broad range of bills, including the Government Bill, Private Members Bill, the Money Bill, the Ordinary Bill, the Financial Bill \& Constitutional Amendment Bill from 1986 to the present. Moreover, on average, there are 3932.99 tokens per sentence. The documents on the proposed dataset are considerably long. They contain richer information of parliament bill content, testing the reasoning abilities and domain-specific capabilities of language models in the legal domain.

    \item \textbf{Quality} Detailed Statistics, analysis of the dataset, and fine-grained evaluation of different parts of documents are provided. Moreover, we also performed extensive quality experiments to evaluate different types of models ranging from RNN to high-performance pre-trained domain models.
    
    \item \textbf{Reproducible Results} We employ the HuggingFace Transformers library \cite{Wolf2019HuggingFacesTS} to facilitate our experiments. Furthermore, we pre-process and publish datasets on HuggingFace Datasets \cite{Lhoest2021DatasetsAC} to reproduce the results and experiment with new models in the future.
    
    \item \textbf{Proposing Two Benchmarks} We are proposing two new benchmarks on the DeepParliament dataset: Binary and Multi-Class Bill Status classification. In addition to the code, we also publish the benchmark on PaperwithCode \footnote{https://paperswithcode.com/dataset/deepparliament} and Open LegalAI \footnote{openlegalai.github.io/DeepParliament} to track the progress.

\end{itemize}
\section{The DeepParliament Dataset}

\subsection{Task Definition}
We model the bill prediction task as a classification problem and design Binary and Multi-Class Classification problem statements on the proposed dataset to evaluate the domain-specific capabilities of language models in the legal domain. For a given collection of labelled bill documents $\mathbf{X}$, the objective is to learn a classification function:
\begin{equation}
    {f : x_{i} \rightarrow {y}_{i}} 
\end{equation}
Where $x_{i}$ is a legal bill document.
\subsection{Task 1: Binary Classification}
In equation (1) $y_{i} \in \{0,1\}$ is target binary label of corresponding status Passed, Failed of classification task.

\subsection{Task 2: Multi-Class Classification}
Task 2, the coarse-grained classification task, had a total of 5 classes. In equation (1) $y_{i} \in \{1,..., K\}$ is the multi-class label of the corresponding status Passed, Negatived, Lapsed, Removed, Withdrawn.
\begin{figure}
\centering
  \includegraphics[width=7.5cm]{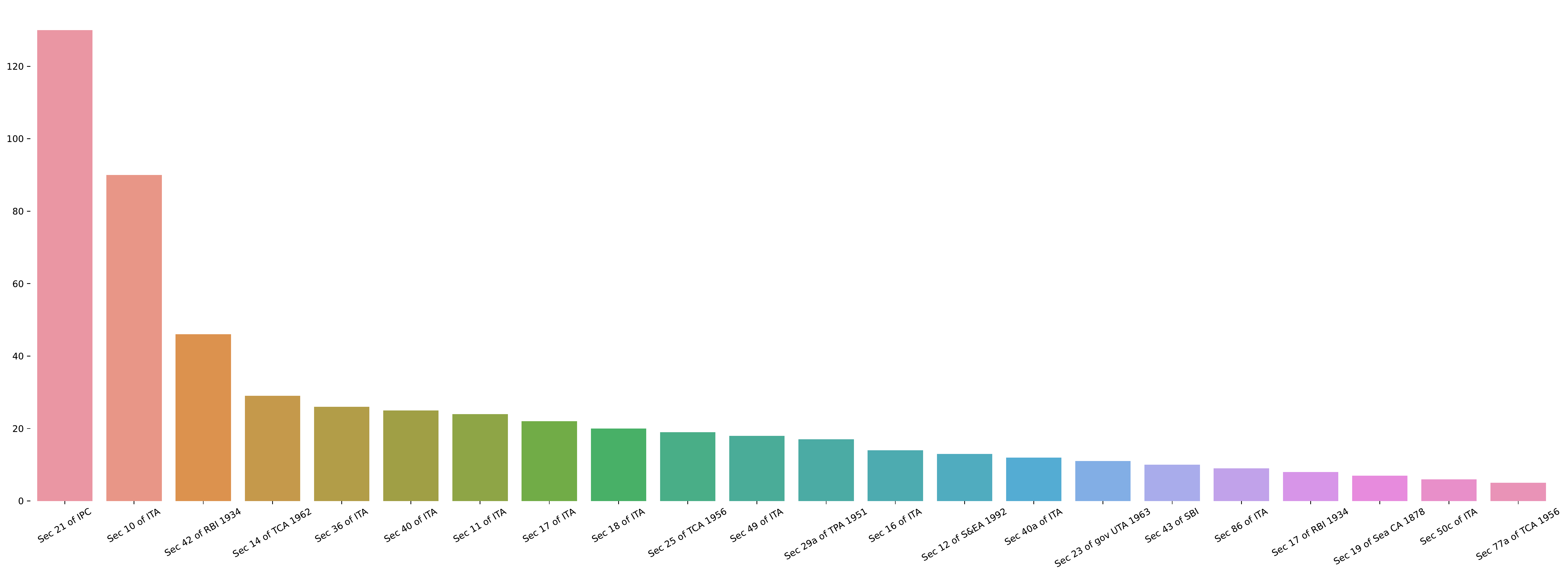}
  \ExerciseCaption{ \footnotesize The Distribution of Top India's Act/Law’s Section mentioned in the dataset. }
  \label{fig:top_sections}
\end{figure}
\subsection{Bill and Lawmaking Procedure}

The foundation of India's democracy is the parliament and state legislatures. Making, changing, and removing laws is the primary responsibility of Parliament. The method by which a legislative proposal is turned into an act is referred to as the legislative process or the lawmaking process in relation to Parliament. The procedure of a new act starts with identifying the need for a new law or an amendment to a current part of the legislation. Following the legal requirement, the relevant ministry writes a text for the proposed legislation, known as a Bill. Other relevant ministries are informed about this bill so that they can make any alterations or amendments.

A bill, which is draft legislation, cannot become law until it has been approved by both houses of Parliament and the President of India. Furthermore, the bill is introduced in Parliament after receiving cabinet approval. Prior to becoming an act, every bill passes through several readings in both houses. After both houses have approved a bill of Parliament, it is forwarded to the president for his or her approval. However, the president can request information and an explanation about the bill. The bill may be returned to Parliament for further consideration. The bill is declared an act with the president's assent. Moreover, the bill is then made into law, and the responsible ministry draughts and submits to Parliament the rules and regulations necessary to carry out the Act.
\section{Dataset Collection \& Preprocessing }
\label{sec:dataset_collection}
We constructed the DeepParliament corpus from raw data collected from the official \footnote{https://loksabhaph.nic.in} \& open website \footnote{https://prsindia.org/billtrack} which put together all the parliament bills from 1986 to the present.  In addition to the raw data, additional metadata is also provided, i.e. the title, type of bill such as government or private, source of the bill, pdf URL and status of the bill. We used pdfminer3 \footnote{https://pypi.org/project/pdfminer3} to extract bill content from each PDF. Some old pdfs are in image format; we applied an OCR system to convert images to text. The pdf content \& metadata were converted into CSV format and combined into a single dataset. Next, we eliminate bill documents with a single token and duplicates. The cleaning pipeline involves removing the special characters, extra spaces etc. All bill documents in this dataset are specifically in the English language.
\section{Dataset statistics}
\label{sec:dataset_statistics}
\begin{figure*}[!ht]
\centering
  \includegraphics[width=13.5cm]{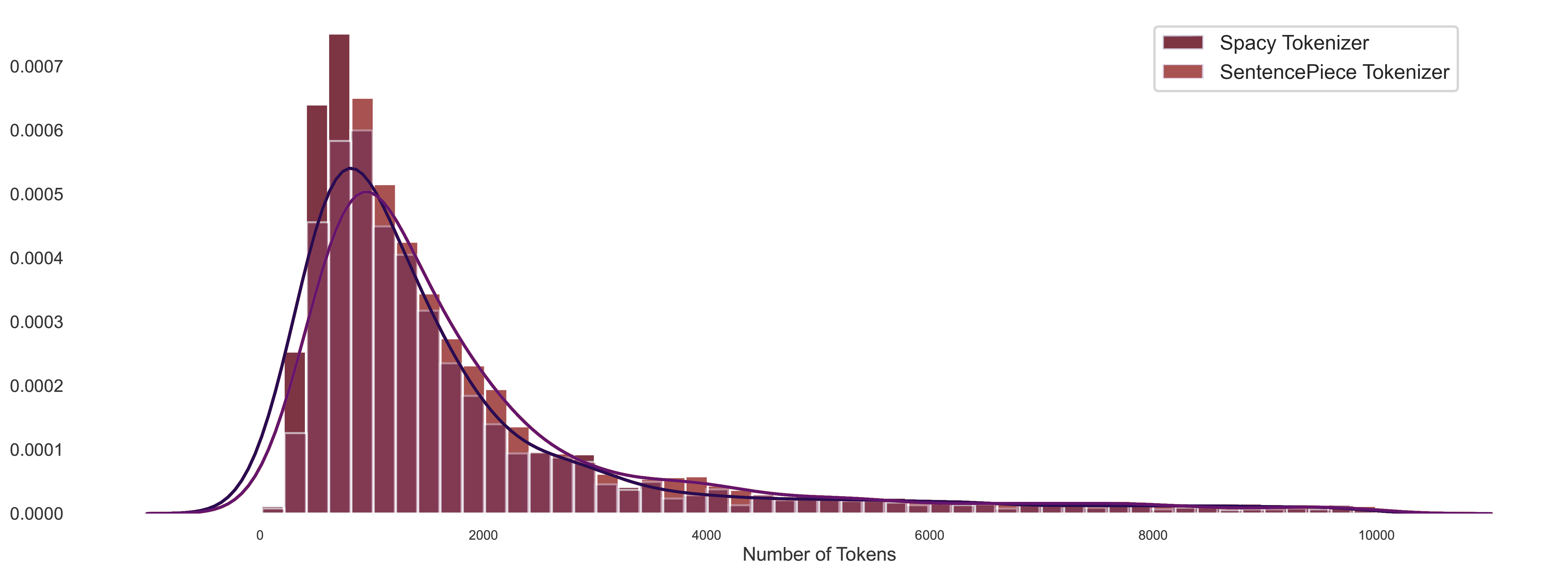}
  \ExerciseCaption{ \footnotesize Distribution of the bill document's length in common words (using the spacy tokenizer) and sub-word units (generated by the SentencePiece tokenizer used in BERT)}
  \label{fig:token_dist}
\end{figure*}

The statistics of our proposed dataset, including the train and the test corpus, are shown in Table \ref{tab:data_stats}. Total documents are 5,329, where 4223 are in the train and 1106 are in the test dataset. Each bill document contains many sentences in both cases, and the document's length varies greatly.
\begin{figure}[!ht]
\centering
  \includegraphics[width=7.5cm]{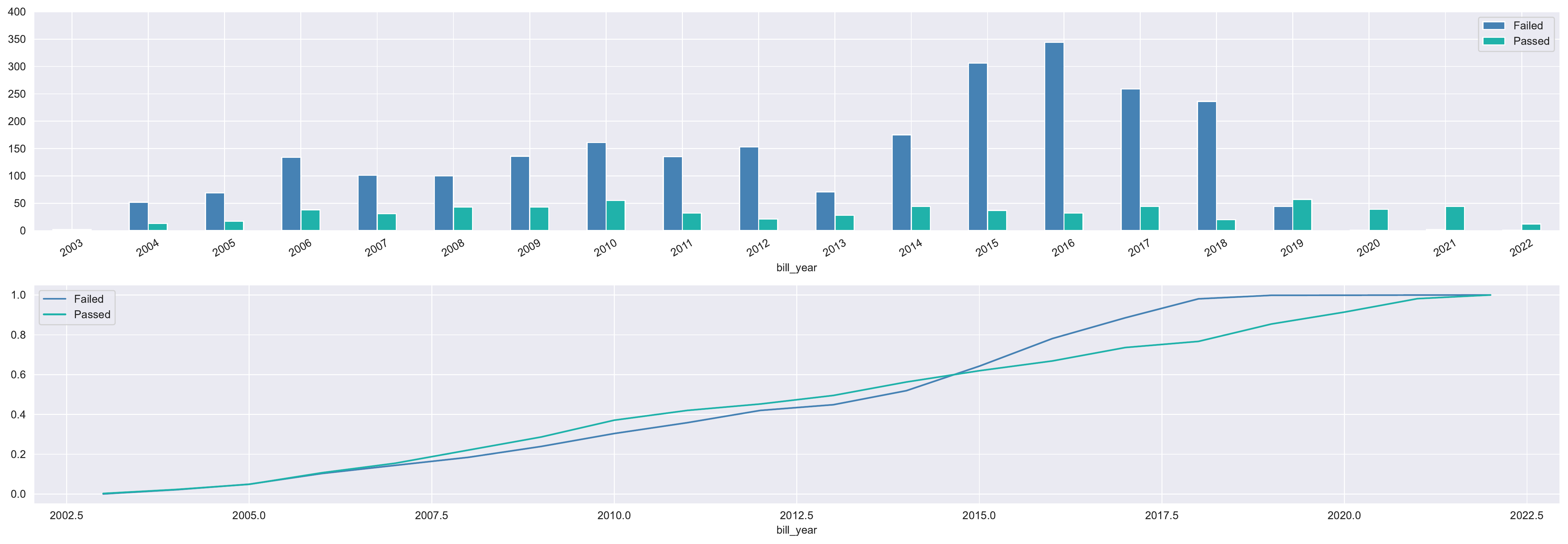}
  \ExerciseCaption{ \footnotesize Distribution \& Cumulative Frequency Graph of Passed and Failed Bill status in the last 20 years.}
  \label{fig:last_20_years}
\end{figure}
The performance of the models is influenced by the amount of vocabulary, which is a good indicator of the linguistic and domain complexity associated with a text corpus; Fig \ref{fig:token_dist} shows the distribution of unique tokens of the train \& test set. As shown in the table, this dataset has 284103 tokens in total, and the train split contains 243393 tokens where the test vocabulary size is 86616. On average, there are 3932.99 tokens per sentence.
\begin{figure}
\centering
  \includegraphics[width=4.5cm]{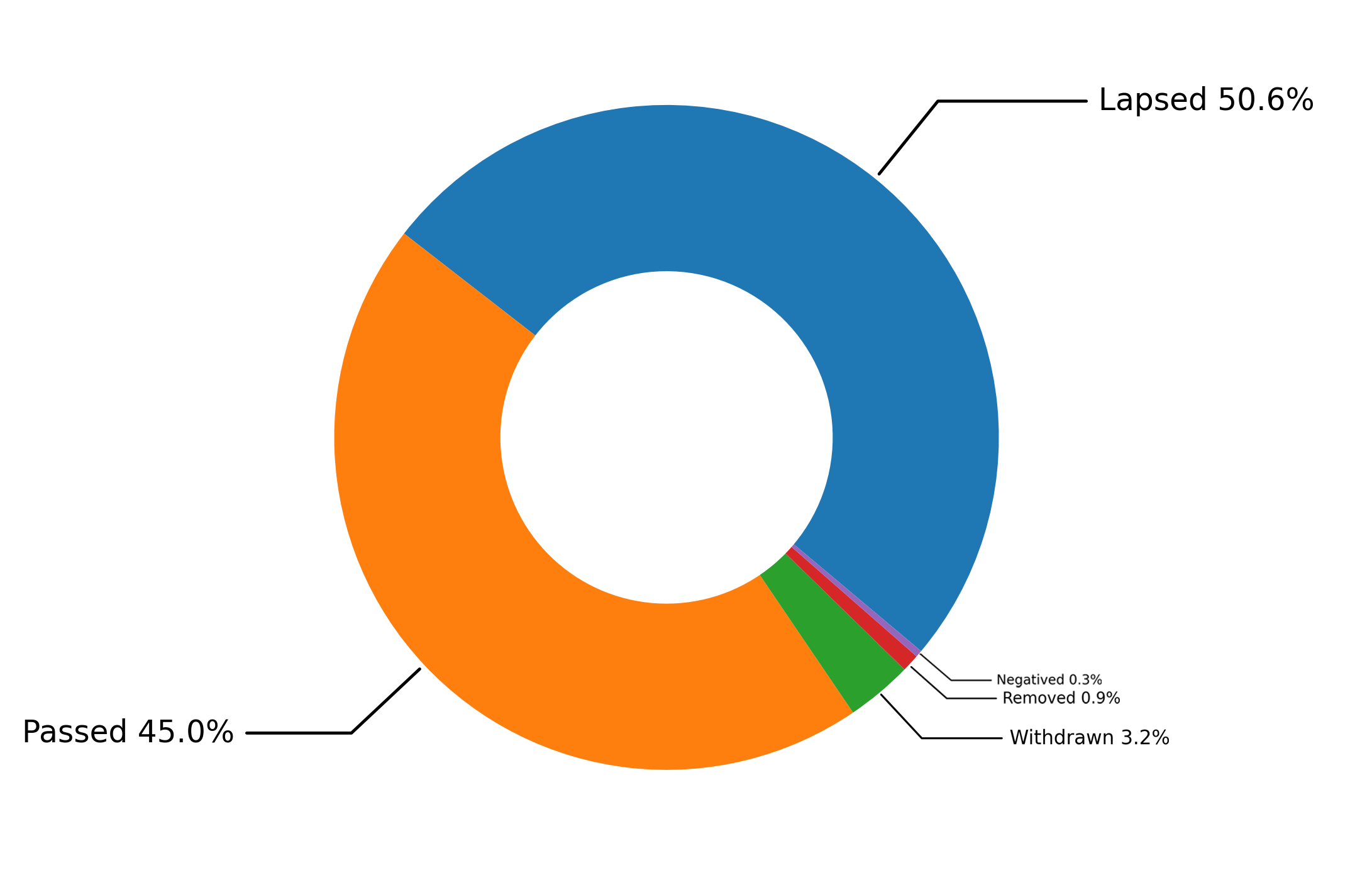}
  \ExerciseCaption{ \footnotesize Relative sizes of documents per bill status in Dataset}
  \label{fig:status_ration}
\end{figure}

\section{Dataset Analysis}
\label{sec:dataset_analysis}

\begin{figure}
\centering
  \includegraphics[width=6.8cm]{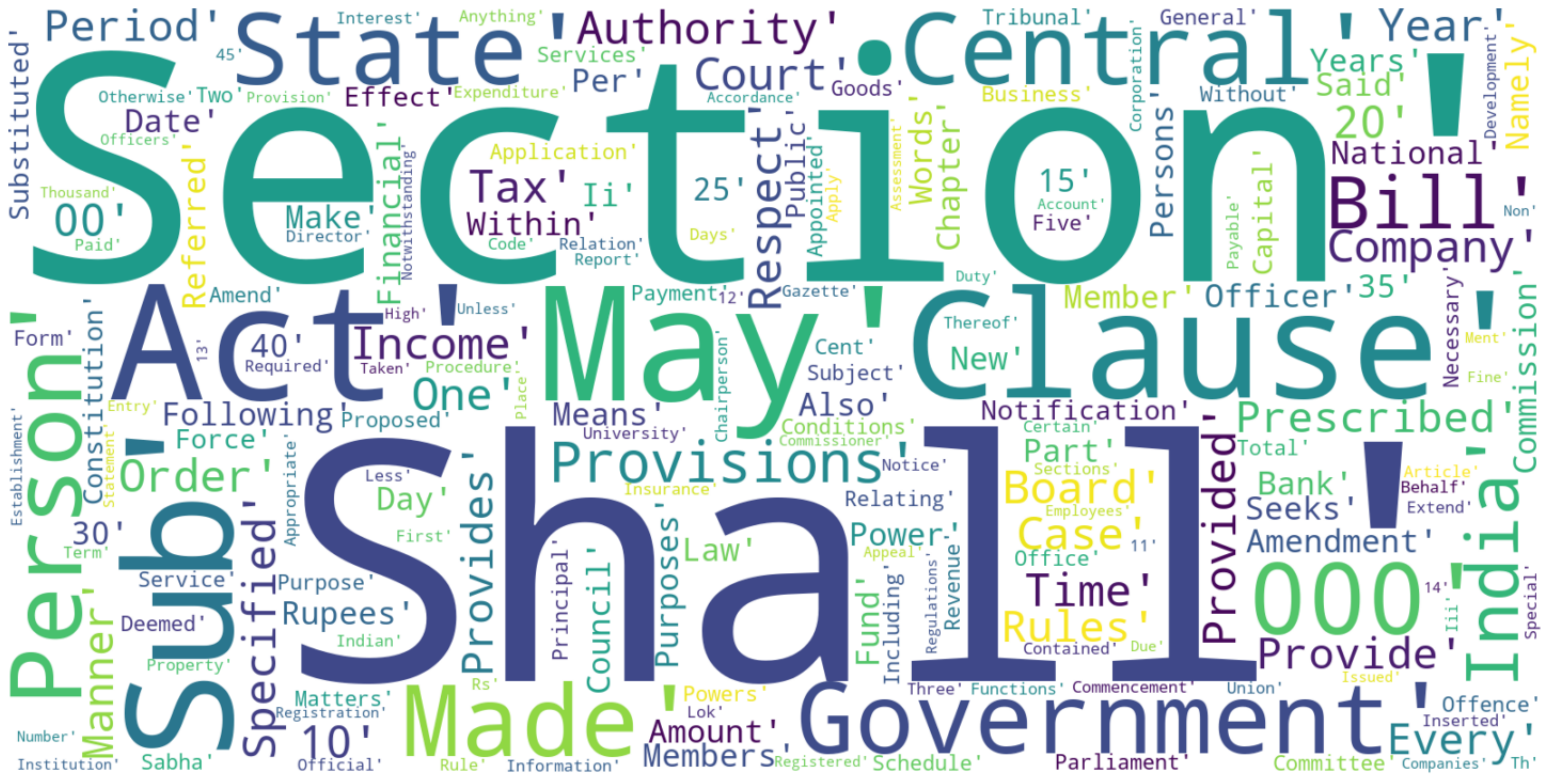}
  \ExerciseCaption{ \footnotesize WordCloud of the top words in the dataset}
  \label{fig:word_cloud}
\end{figure}

The documents on the proposed dataset are considerable long and contain richer information on bill content. As described before, the dataset has been categorized into two settings: Binary Classification and Multi-Class classification. The most frequent category status is Lapsed, which occupies 50.6\%. Fig \ref{fig:status_ration} shows the percentage of each status type. Lapsed, Passed, and Withdrawal is the dataset's top three common statuses. The proposed dataset text covers a broad range of bills, including Government Bill, Private Member Bill, Money Bill, Ordinary Bill, Financial Bill and Constitutional Amendment Bill.
We used word cloud to visualize the top Commonly occurring legal words in the dataset shown in Fig \ref{fig:word_cloud}.

We visualized the top sections mentioned in the entire dataset. A section is a specific provision of a legal code or body of laws, often laying out a specific legal obligation. Most sections in the dataset come under the Indian panel code and income tax act. Fig. \ref{fig:top_sections} shows the top Indian Act/Law’s Section mentioned in the dataset. To understand the dataset better, we also visualize the bill status of the last top 20 years. Fig. \ref{fig:last_20_years} shows the visualization. We can see that year 2015 and 2016 has the most significant failure ratio in the last 20 years, while in 2019, most bills were passed compared to other years.

\begin{table}[!ht]
\footnotesize
\centering
\resizebox{0.40\textwidth}{!}{%
\begin{tabular}{lcccc}
\toprule
 & {\bf Train} & {\bf Test} & {\bf Total} \\
\midrule
Documents \#  &  4223 & 1106  & 5329 \\
Vocab & 243393& 86616 & 284103  \\
Max D tokens & 219378 & 227407 & 227407   \\
Max T tokens & 36 & 36& 36  \\
Avg D tokens & 3932.99 & 4080.97 & 3963.70  \\
Avg T tokens & 11.15 & 11.48 & 11.22 \\
\bottomrule
\end{tabular}}
\ExerciseCaption{\footnotesize DeepParliament dataset statistics, where D, T represents the Documents and Title, respectively}
\label{tab:data_stats}
\vspace{-2ex}
\end{table}

\section{Methods}
Our study considers ten text classification models ranging from long-range RNN and CNN to Transformer-based methods increasing recency and sophistication.
\subsection{Sequence \& Convolutional models}
In this category, we experimented with standard long short-term memory (LSTM) \cite{Hochreiter1997}, Bidirectional LSTM (BiLSTM) and convolutional neural network (CNN) \cite{Kim2014} models for bill prediction tasks. 
We decided to use a shallow CNN model for glove since research has shown that deep CNN models do not consistently outperform other algorithms for text classification tasks. Initially, we utilized Xavier weight initialization \cite{Glorot2010} for both models' embedding matrices. Later we leverage this by initializing word vectors using pre-trained GloVe embedding \cite{Pennington2014} of length 300.
\subsection{General Domain Pre-trained Models}
We experiment with Transformer based model BERT \cite{Devlin2019} and its variants. 
RoBERTa \cite{Liu2019} and ALBERT \cite{Lan2020} are an extension of the standard BERT model. RoBERTa uses a byte-level BPE as a tokenizer and a different pre-training scheme where ALBERT's model has a smaller parameter size than corresponding BERT models.
\subsection{Legal Domain Pre-trained Models}
Recent research has also demonstrated that language representation models trained on massive corpora and precisely adjusted for a particular domain task perform much better than models trained on task-specific data. This method of transfer learning is beneficial in legal NLP. 
We thus evaluated three Pre-trained Language models trained from scratch with legal documents, including Legal-BERT \cite{Chalkidis2020}, Legal-RoBERTa and Custom Legal-BERT \cite{Zheng2021}.

\section{Experiments}
In this section, we evaluate the mentioned models on the proposed Dataset, describe the executed experiments, and examine the results.
\subsection{Experimental Settings}
In all sequence models, the batch size was set to 64, and the number of epochs was set to 50. At the same time, we iterate through 50 epochs with a batch size of 8 for all Bert-based models.
\begin{table*}[t!]
    \centering
    \resizebox{0.85\textwidth}{!}{%
    \begin{tabular}{@{}l|c@{ }c@{ }c@{ }|c@{ }c@{ }c@{ }|ccc@{ }}
    \toprule
    {} & \multicolumn{3}{c|}{\bf Starting Tokens} & \multicolumn{3}{c|}{\bf Middle Tokens} & \multicolumn{3}{c}{\bf End Tokens}\\
    \midrule
    \bf Model & \multicolumn{1}{c}{\bf Precision} &  \multicolumn{1}{c}{\bf Recall} & \multicolumn{1}{r|}{\bf F1-score} & \multicolumn{1}{c}{\bf Precision} & \multicolumn{1}{c}{\bf Recall} & \multicolumn{1}{r|}{\bf F1-score} & \multicolumn{1}{c}{\bf Precision} &  \multicolumn{1}{c}{\bf Recall} &  \multicolumn{1}{c}{\bf F1-score} \\
    \midrule
   
   ALBERT$_{Base}$ & 92.21 & 92.54 & 92.28 & 90.13 & 90.38 & 90.19 &                 89.06 &              89.31 &          88.08 \\
Custom Legal-BERT$_{Base}$ &                  92.47 &               92.83 &           92.47 &                   89.51 &                89.80 &            89.56 &                 89.03 &              89.31 &          89.10 \\
RoBERTa$_{Base}$ &                  92.74 &               93.06 &           92.83 &                   89.30 &                89.53 &            89.36 &                 89.42 &              89.70 &          89.47 \\
Legal-RoBERTa$_{Base}$ &                  92.89 &               93.17 &           92.92 &                   90.24 &                90.51 &            90.32 &                 90.42 &              90.63 &          90.45 \\
Bert$_{Base}$ &                  92.92 &               93.23 &           93.01 &                   \textbf{90.68} &                \textbf{90.95} &            \textbf{90.73} &                 90.02 &              90.06 &          90.19 \\
Legal-BERT$_{Base}$ &                  \textbf{93.11} &               \textbf{93.49} &           \textbf{93.11} &                   90.62 &                90.93 &            90.61 &                 \textbf{90.42} &              \textbf{90.64} &          \textbf{90.49} \\

    \bottomrule
    \end{tabular}
    }
    \ExerciseCaption{\footnotesize Performance of all Transformer based baseline models in Macro-Precision, \captionlinebreak Macro-Recall and Macro-F1 (\%) on Binary test set under different tokens settings.
    }
    \label{tab:accbinary_token}
\end{table*}

\begin{table*}[t!]
    \centering
    \resizebox{0.85\textwidth}{!}{%
    \begin{tabular}{@{}l|c@{ }c@{ }c@{ }|c@{ }c@{ }c@{ }|ccc@{ }}
    \toprule
    {} & \multicolumn{3}{c|}{\bf Starting Tokens} & \multicolumn{3}{c|}{\bf Middle Tokens} & \multicolumn{3}{c}{\bf End Tokens}\\
    \midrule
    \bf Model & \multicolumn{1}{c}{\bf Precision} &  \multicolumn{1}{c}{\bf Recall} & \multicolumn{1}{r|}{\bf F1-score} & \multicolumn{1}{c}{\bf Precision} & \multicolumn{1}{c}{\bf Recall} & \multicolumn{1}{r|}{\bf F1-score} & \multicolumn{1}{c}{\bf Precision} &  \multicolumn{1}{c}{\bf Recall} &  \multicolumn{1}{c}{\bf F1-score} \\
    \midrule
Custom Legal-BERT$_{Base}$ &                  51.49 &               49.12 &           49.42 &                   42.79 &                39.96 &            40.58 &                 41.10 &              41.21 &          43.25 \\
ALBERT$_{Base}$ &                  56.14 &               51.55 &           52.87 &                   52.55 &                46.28 &            47.73 &                 50.71 &              45.05 &          45.99 \\
RoBERTa$_{Base}$ &                  56.07 &               54.85 &           54.89 &                   49.00 &                47.05 &            47.60 &                 45.57 &              45.11 &          45.03 \\

Legal-RoBERTa$_{Base}$ &                  61.40 &               54.74 &           57.44 &                   48.08 &                44.02 &            45.53 &                 50.89 &              46.98 &          49.06 \\
Bert$_{Base}$ &                  60.55 &               55.92 &           57.86 &    \textbf{63.75} &                \textbf{52.61} &            \textbf{55.64} &                 46.13 &              46.17 &          46.34 \\

Legal-BERT$_{Base}$ &                  \textbf{62.96} &               \textbf{56.96} &           \textbf{58.79} &                   55.47 &                49.15 &            49.86 &                 \textbf{53.50} &              \textbf{47.20} &          \textbf{49.68} \\
    \bottomrule
    \end{tabular}
    }
    \ExerciseCaption{ \footnotesize Performance of all Transformer based baseline models in Macro-Precision, \captionlinebreak Macro-Recall and Macro-F1 (\%) on Multi-Class test set under different tokens settings.
    }
    \label{tab:accmulti_token}
\end{table*}

\begin{figure*}[!ht]
\centering
  \includegraphics[width=13.5 cm]{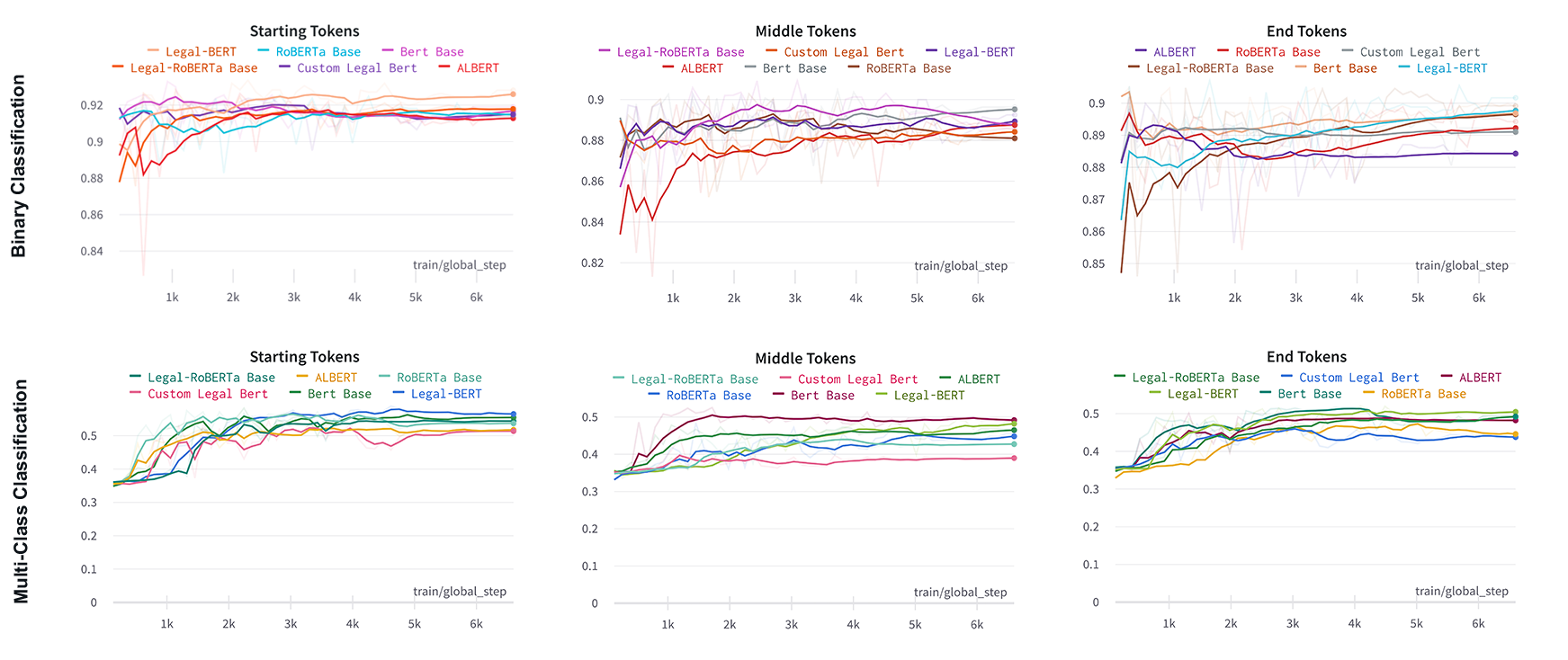}
     \ExerciseCaption{ \footnotesize Macro-F1 scores of different Transformer based models on the test dataset.}
  \label{fig:tokenaccplot}
\end{figure*}
We used Tensorflow's Keras API \cite{tensorflow} to build sequence models. The BERT-based model follows the base configuration, consisting of 12 layers, 786 units, and 12 attention heads. We developed these models using Pytorch \cite{NEURIPS2019_9015} and obtained pretrained checkpoints from the HuggingFace library \cite{Wolf2019HuggingFacesTS}.

We evaluate the models in two settings: Binary Classification and Multi-Class Classification. For the first setting, the classification layer consists of a dense layer with 1 unit as output, with sigmoid activation. The loss was calculated using binary cross-entropy. 

\begin{equation}
\mathrm{Loss}_{\mathrm{bce}} = -\frac{1}{N} \sum_{n=1}^{N}\left[y_{n} \log \hat{y}_{n}+\left(1-y_{n}\right) \log \left(1-\hat{y}_{n}\right)\right]
\end{equation}
In The Multi-Class setting, we used a dense layer with five units as output, with softmax activation. In this case, categorical cross-entropy was used for loss calculation.

\begin{equation}
\centering
\mathrm{Loss}_{\mathrm{cce}} =-\sum_{i=1}^{\operatorname{N}} y_{i} \cdot \log \hat{y}_{i}
\end{equation}
We perform five runs with different seeds for each method and report the average scores. All the experiments were conducted on the Google Colab Pro and used the default GPU Tesla T4 16GB. The proposed dataset \& code are available at github.com/monk1337/DeepParliament for reproducibility.

\subsection{Evaluation Metrics}
We assessed the baseline and other models based on their Macro-averaged F1 scores, accuracy, and recall in multi-class and Binary environments. Before calculating the average across labels, macro-averaging computes the metric inside each label.

\section{Results \& Discussion}

Table \ref{tab:full_binary} \& Table \ref{tab:full_multi} shows the performance of all models in Macro-Precision, Macro-Recall and Macro-F1 on Binary \& Multi-Class test set respectively under full token settings.
Under the Sequence \& Convolutional models category, LSTM performed better than Vanilla CNN in both the Binary and Multi-Class Bill Prediction tasks.

It is observed that there is a significant improvement in the model's performance when Glove embedding is used as word vectors compared to other embeddings results. BiLSTM + Glove performed best in sequential and convolutional models. CNN + glove gave the second-best results in this category. In the General Domain of Pre-trained Models, transformer models outperform sequential \& Convolutional models.
Bert$_{Base}$ performed best in both Binary and Multi-Class settings while RoBERTa$_{Base}$ and ALBERT$_{Base}$ are a close second with better f1-score of all the models based in Multi-Class and Binary settings, respectively.

Our assumption was that legal domain models would not perform well on the proposed dataset as India's legal systems are completely different. However, our assumption did not hold true. In a few settings, Domain-specific models performed well compared to general domain models; This is likely because most words in the proposed dataset are legal domain-specific. The high frequency of unique, domain-specific terminologies appears in the dataset but not in the vocabulary of the Transformer Models trained on the general domain. It is observed that Bert$_{Base}$ performs best in terms of precision, while In legal domain trained models, the Legal-BERT$_{Base}$ model performs better in recall and f1 score in the Multi-Class classification task. On the other hand, in the Binary classification task, Legal-RoBERTa$_{Base}$ \& Custom Legal-BERT$_{Base}$ performs better than other models.
\begin{table}[t!]
\small
\centering
    \resizebox{0.40\textwidth}{!}{%
    \begin{tabular}{@{}l|c@{ }|c@{ }|cc@{ }}
    \midrule
    \bf Model & {\bf Precision} &  {\bf Recall} & {\bf  F1-score} \\
    \midrule
    CNN  &72.8&57.8&47.1\\
    LSTM &57.2&57.0&57.0\\
    CNN + Glove  &71.6&67.6&64.4\\
    BiLSTM + Glove &66.4&66.0&64.9\\
    ALBERT$_{Base}$  &91.7&92.1&91.7\\
    RoBERTa$_{Base}$  &92.2&92.5&92.2\\
    Bert$_{Base}$  &92.4&92.7&92.5\\
    Legal-BERT$_{Base}$  &92.7&93.0&92.7\\
    Custom Legal-BERT$_{Base}$  &92.8&93.1&92.7\\
    \midrule
    \rowcolor{rowgray}
    \textbf{Legal-RoBERTa$_{Base}$}  &93.1&93.4&93.1\\
    \bottomrule
    \end{tabular}
    }
    \ExerciseCaption{\footnotesize Performance of all baseline models in Macro-Precision, Macro-Recall and Macro-F1 (\%) on Binary test set under full tokens setting.
    }
    \label{tab:full_binary}
\end{table}
\subsection{Which portions of the bill contain the most useful information?}
Legal documents are lengthy and include specialist terminology compared to conventional corpora used to train text classification and language models. We did not employ the Longformer \cite{Beltagy2020} and Reformer \cite{Kitaev2020} models explicitly designed for lengthy texts due to memory and GPU constraints. We initially experimented with the pre-trained models trained on general-purpose texts. We experimented with various portions of the documents, including Starting tokens, Middle tokens and End tokens, to overcome the restriction on the number of input tokens Bert and other transformer models accept. Table \ref{tab:accbinary_token} \& Table \ref{tab:accmulti_token} shows the Performance of all Transformer based models in Macro-Precision, Macro-Recall and Macro-F1 on Binary \& Multi-Class test set respectively under different token settings. 

Among all the combinations of input tokens, we observed that the performance of the prediction algorithm improves when more tokens from the first and middle document sections are being used as input. This leads us to infer that the first and middle portion of the documents contains the most helpful information. The proposed dataset shares the quality of including many domain-specific words relevant to the law. When the dataset is limited, the models depend on prior knowledge utilizing the transfer learning. Legal-BERT$_{Base}$ uncased produced the highest macro-averaged F1 score across first and last token settings under both the Multi-Class \& Binary classification categories. Legal-BERT's prior learning is more applicable to the proposed benchmarks.
\begin{table}[t!]
\small
\centering
    \resizebox{0.40\textwidth}{!}{%
    \begin{tabular}{@{}l|c@{ }|c@{ }|cc@{ }}
    \midrule
    \bf Model & {\bf Precision} &  {\bf Recall} & {\bf  F1-score} \\
    \midrule
    CNN  &26.7&21.3&15.3\\          
    LSTM &19.8&19.6&18.3\\
    CNN + Glove  &25.2&22.6&18.8\\
    BiLSTM + Glove &27.4&27.3&26.6\\
    RoBERTa$_{Base}$  &60.0&43.4&45.3\\
    ALBERT$_{Base}$  &52.7&46.3&47.6\\
    Custom Legal-BERT$_{Base}$  &54.0&54.5&53.8\\
    Legal-RoBERTa$_{Base}$  &58.1&56.8&57.1\\
    Bert$_{Base}$   &65.2&54.6&58.0\\
    \midrule
    \rowcolor{rowgray}
    \textbf{Legal-BERT$_{Base}$}  &64.9&59.3&61.4\\
    \bottomrule
    \end{tabular}
    }
    \ExerciseCaption{\footnotesize Performance of all baseline models in Macro-Precision, Macro-Recall and Macro-F1 (\%) on Multi-Class test set under full tokens setting.
    }
    \label{tab:full_multi}
\end{table}

Moreover, Bert$_{Base}$ performed well in the middle token setting in both classification categories. At the same time, Legal-RoBERTa$_{Base}$ emerged as the second best performing model under the last token setting. Figure \ref{fig:tokenaccplot} shows the visualization of Macro-F1 scores of different models on the test dataset. We discovered that the best model had a significant advantage over the general domain model since it had been pre-trained in the same language and on data specific to the domain.

\section{Conclusion}
In this work, DeepParliament, A Legal domain Benchmark Dataset, is presented, which requires a deeper domain and language understanding in the legal field. It covers a broad range of parliament bills from 1986 to the present and tests the reasoning abilities of a model. 
Based on Extensive quality experiments on different models, It is shown that the dataset is a challenge to the present state-of-the-art methodologies and domain-specific models, with the best baseline obtaining just 59.79\% accuracy. This dataset is anticipated to aid future studies in this field.

\section*{Limitations}
DeepParliament is limited to evaluating English models at this time. In India, bill documents are also available in other local languages. Developing models \& datasets for other languages would be an essential road for future research. Besides language, the current version of DeepParliament is also limited by size. However, we will continue to prioritize adding new bill documents from official sources; introduced in either house of Parliament, i.e. the Lok Sabha or the Rajya Sabha. Documents in the dataset are long and unstructured. Current Transformers models are limited by their input size and cannot process full documents at once. Extended sequence models such as Longformer \cite{Beltagy2020} and BigBird \cite{zaheer2020bigbird} are not currently evaluated on the dataset. We leave the investigation of those models on the proposed dataset for other groups to experiment with and publish the results.

Despite the limitations as mentioned above, we believe that the dataset will be helpful to many researchers. as it takes the initial steps to establish a well-defined benchmark to evaluate legal domain models in this field. Models developed on this dataset may assist MPs, presidents, and other legal practitioners.

\section*{Ethics Statement}

This study focuses on proposing the first dataset on Parliament Bill status prediction, adheres to the ethical guidelines outlined in the ACL code of Ethics and examines the ethical implications.
DeepParliament gathers its data from two public sources. There is no privacy concern since all bill documents are collected against open-access databases. Moreover, the documents do not include personal or sensitive information, except minor information provided by authorities, such as the names of the presidents, Union Council of Ministers, and other official administrative organisations.

The details of dataset collection and statistics are provided in Sections \ref{sec:dataset_collection}
and \ref{sec:dataset_statistics}. The model trained to utilise our dataset is mainly meant to support decision-making during bill analysis, not to replace the human specialists.

\bibliography{anthology,custom}
\end{document}